\documentclass[sn-nature]{sn-jnl}% Style for submissions to Nature Portfolio journals
\usepackage{amsmath}
\usepackage{graphicx}%
\usepackage{multirow}%
\usepackage{amsmath,amssymb,amsfonts}%
\usepackage{amsthm}%
\usepackage{mathrsfs}%
\usepackage[title]{appendix}%
\usepackage{xcolor}%
\usepackage{textcomp}%
\usepackage{manyfoot}%
\usepackage{booktabs}%
\usepackage{algorithm}%
\usepackage{algorithmicx}%
\usepackage{algpseudocode}%
\usepackage{listings}%
\usepackage{caption}
\usepackage[nolists,nomarkers]{endfloat}
\usepackage{lineno}

\theoremstyle{thmstyleone}%
\theoremstyle{thmstyletwo}%

\theoremstyle{thmstylethree}%

\begin{document}
% \linenumbers 
\title[Article Title]{Harnessing Deep Learning of Point Clouds for Inverse Control of 3D Shape Morphing}


\author[1]{\fnm{Jue} \sur{Wang}}\email{wang5056@purdue.edu}

\author[1]{\fnm{Dhirodaatto} \sur{Sarkar}}\email{sarkar40@purdue.edu}

\author[2]{\fnm{Jiaqi} \sur{Suo}}\email{suoj@purdue.edu}

\author*[1]{\fnm{Alex} \sur{Chortos}}\email{achortos@purdue.edu}

\affil*[1]{\orgdiv{School of Mechanical Engineering}, \orgname{Purdue University}, \orgaddress{\street{610 Purdue Mall}, \city{West Lafayette}, \postcode{47907}, \state{Indiana}, \country{USA}}}

\affil[2]{\orgdiv{Schools of Construction Management Technology}, \orgname{Purdue University}, \orgname{Purdue University}, \orgaddress{\street{610 Purdue Mall}, \city{West Lafayette}, \postcode{47907}, \state{Indiana}, \country{USA}}}


\abstract{Shape-morphing devices, a crucial branch in soft robotics, hold significant application value in areas like human-machine interfaces, biomimetic robotics, and tools for interacting with biological systems. To achieve three-dimensional (3D) programmable shape morphing (PSM), the deployment of array-based actuators is essential. However, a critical knowledge gap impeding the development of 3D PSM is the challenge of controlling the complex systems formed by these soft actuator arrays. This study introduces a novel approach, for the first time, representing the configuration of shape morphing devices using point cloud data and employing deep learning to map these configurations to control inputs. We propose Shape Morphing Net (SMNet), a method that realizes the regression from point cloud data to high-dimensional continuous vectors. Applied to previous 2D PSM actuator arrays, SMNet significantly enhances control precision from $82.23\% $to $97.68\%$. Further, we extend its application to 3D PSM devices with three different actuator mechanisms, demonstrating the universal applicability of SMNet to the control of 3D shape morphing technologies. In our demonstrations, we confirm the efficacy of inverse control, where 3D PSM devices successfully replicate target shapes. These shapes are obtained either through 3D scanning of physical objects or via 3D modeling software. The results show that within the deformable range of 3D PSM devices, accurate reproduction of the desired shapes is achievable. The findings of this research represent a substantial advancement in soft robotics, particularly for applications demanding intricate 3D shape transformations, and establish a foundational framework for future developments in the field.}




\maketitle

\section{Introduction}\label{sec1}

Shape-morphing devices, a pivotal subset of soft robotics, aspire to achieve programmable, controllable, and reversible transformations reminiscent of biological systems such as octopi and growing plants. They exhibit potential in realms such as human-machine interfaces for augmented and virtual reality (AR/VR) devices\cite{klemmer2006bodies}, haptic technology\cite{yu2019skin}, optical and acoustic metamaterials\cite{peng2020dynamically}, and devices for manipulating biology.\cite{chen2022harnessing, kirillova2019shape, viola2020guiding} 

The most commonly investigated shape morphing devices morph between two distinct shapes, driven by the material design or structural configurations.\cite{ford2019multifunctional,kotikian20183d,coelho2008surflex,yu2013electronically,nojoomi2018bioinspired,mao20163d,wu2016multi} We refer to these devices as pattern-to-pattern shape morphing (PPSM). Many emerging applications require controllable and reversible transformations, which has stimulated the development of devices that can transform their structure on demand, which we refer to as programmable shape morphing (PSM)\cite{coelho2008surflex, stanley2016closed,  liu2021robotic, ni2022soft, rauf2023electroadhesive, bai2022dynamically, wang2023passively}. Such devices consist of an array of actuators, enabling a singular device to transform into various configurations as necessitated. 

Early PSM systems were composed of arrays of solid linear actuators that could reproduce surfaces on demand.\cite{hirota1995surface,iwata2001project} Since all actuators were mechanically decoupled, the control algorithms were relatively simple. However, the bulky and cumbersome nature of the linear actuators and their control equipment\cite{leithinger2010relief,follmer2013inform}  limited their applications.

Recent advancements in materials and fabrication techniques have led to the emergence of flexible actuators, substantially reducing the size of actuator arrays and enabling the creation of entirely continuous surfaces. This has revitalized interest in PSM. Given the intricate coupling that exists between the deformations of different actuators, the control algorithm plays an important role in continuous PSMs. For the works whose deformation is generated by rod-shaped actuators connected by points\cite{bai2022dynamically, liu2021robotic}, the construction of analytical models is feasible, albeit necessitating certain simplifications. However, the reliance on simplifying assumptions restricts the design freedom of devices. Continuum actuators that actuate throughout their surface or volume\cite{stanley2016closed,wang2021design,wang2023passively} present a more generalized approach to deformation and mimic the continuous nature of biological systems. Yet, the pronounced geometric coupling inherent to arrays of continuous actuators presents significant hurdles for traditional analytical models. Machine learning has recently emerged as a strategy to achieve model-free control of these complex systems.\cite{wang2021design,wang2023passively}

Currently, research on shape morphing devices has predominantly centered around 2D arrays of actuators that are capable of transforming into 3D surfaces\cite{bai2022dynamically, wang2023passively, hajiesmaili2022programmed} due to the availability of 2D fabrication approaches. With the increasing development of 3D fabrication techniques\cite{chortos2021printing, kotikian2021innervated}, 3D arrays of actuators have become viable. However, for controlling these 3D arrays of coupled actuators, their complexity increases exponentially compared to the existing 2D arrays. Consequently, the control algorithms are the crucial knowledge gap in achieving 3D shape morphing.

For soft robots with serial structures, it is possible to use parametric representation of the deformed geometry to accomplish machine learning-based model-free control\cite{jiang2017two,reinhart2016hybrid}. For shape morphing devices with 2D arrays of actuators, one-dimensional data is sufficient to describe their deformation.\cite{bai2022dynamically, liu2021robotic，wang2021design,wang2023passively} However, for the highly complex 3D arrays of actuators, neither parametric representation nor one-dimensional data can adequately capture their intricate deformations in 3D space. Therefore, point cloud data is the most direct and specific method for representing these deformed geometries. 

The versatility of point clouds is reflected by their widespread adoption in fields that rely on 3D representations, including the construction industry\cite{xu2021voxel}, autonomous navigation\cite{zeng2018rt3d}, computer vision\cite{guo2020deep}, and robot sensing\cite{duan2021robotics,pomerleau2015review}. Building upon this, the application of machine learning to point cloud data has opened new frontiers. Machine learning tasks for point clouds primarily encompass segmentation\cite{zhang2019review,xie2020linking}, classification\cite{grilli2017review, zhang2023deep}, and reconstruction.\cite{berger2014state, ma2018review} These methodologies have found prolific applications in areas such as robotic sensing, autonomous driving, and geoscience. While regression tasks with point clouds are less prevalent, they have garnered attention in niche domains including forest biomass estimation\cite{oehmcke2022deep}, reconstruction of deformable objects\cite{lv2023learning}, and hand pose recognition\cite{chen2018shpr}. Inverse control of 3D shape morphing, which involves controlling their deformation based on inputted target shapes, can be described as a mapping between the target shape, expressed in the point cloud, and the control inputs. It can be considered a typical point cloud regression task. 

Therefore, in this study, we present a universal approach to inversely control 3D shape morphing devices with different actuation principles. For the first time, we express the deformation of shape morphing devices using 3D point cloud data and employ deep learning tools to correlate this representation with the input vector that represents the control inputs to an array of actuators. (Fig. 1a) This methodology facilitates the calculation of actuator control signals based on a given target shape, enabling the actuator to achieve the desired configuration. Our training data is derived from Finite Element Analysis (FEA) simulations. We initiated our research by establishing a simulation model for the 3D shape morphing device. Subsequently, random inputs were generated to procure the deformed 3D point cloud data from the simulations. Here, we proposed a new training architecture, named Shape Morphing Net (SMNet), to map the point cloud data with continuous high-dimensional input vectors. This model not only significantly enhances the control precision of 2D low-profile Programmable Shape Morphing (PSM) but also expands its application to its 3D counterpart. The general applicability of this model-free control further allows its extension to 3D PSMs composed of actuators based on two distinct deformation mechanisms. (Fig. 1b-d) To validate the efficacy of our control method in replicating real-world objects, we captured point cloud data of a physical object using a 3D scanner. This data was processed and fed into a pre-trained model, resulting in a control input array for 3D shape morphing. Putting these control inputs into the finite element model replicated the shape of the object. (Fig. 1e) Additionally, we demonstrated that our proposed method is equally capable of reproducing the target shapes derived from more complex virtual 3D models.

\begin{figure}
\centering
\includegraphics[width=14cm]{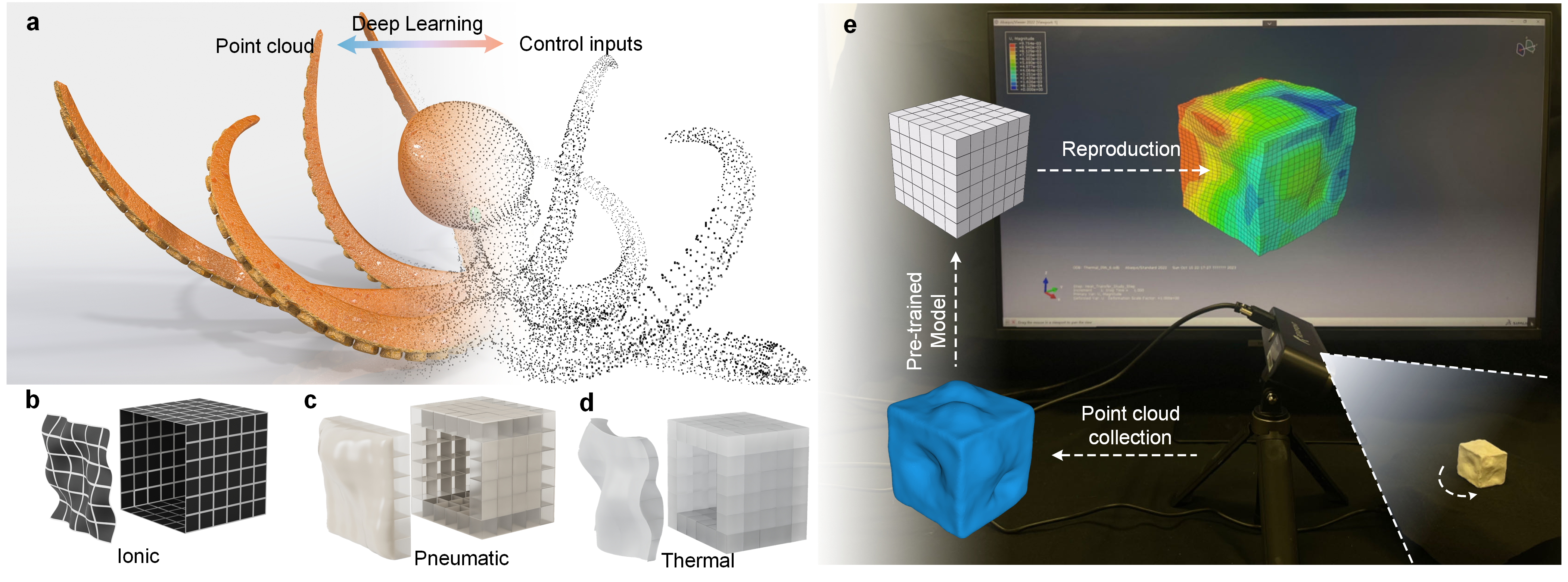}
\caption{\textbf{A universal method for controlling 3D shape morphing devices by mapping the point cloud of the deformed configuration with the control inputs of devices.} a, Utilizing the point cloud to express the configurations of shape morphing devices, using an octopus to symbolize a shape morphing device. b, The rendering for 3D shape morphing devices based on ionic actuator arrays. c, The rendering for 3D shape morphing devices based on thermal actuator arrays. d, The rendering for 3D shape morphing devices based on pneumatic actuator arrays. e, the demonstration procedures by which 3D shape morphing devices reproduce the shape of physical objects in simulations.}
\label{fig::figure1}
\end{figure}

\section{Results}\label{sec2}

\subsection{Data collection and preprocessing of point cloud data}\label{sec3}

In this study, we employed four datasets for training. One dataset originates from our group's prior work,\cite{wang2023passively} on continuous actuator arrays made of ionic actuators. The other three datasets are created from 3D PSMs with three different actuation mechanisms.  All data are derived from FEA simulations, with detailed simulation processes delineated in the Methods section. Post-simulation, we extracted the "XYZ" displacement data for nodes, encompassing two segments: pre-deformation node positions and post-deformation node displacements. Summing them yielded the 3D point cloud data for the post-deformation state. Given the variability in mesh and 3D models across simulations, we implemented a standardized preprocessing routine for the point clouds.
The process of reproducing a 3D structure begins with a 3D scan (Fig 1e) that provides the external points of the structure. Consequently, the first step of preprocessing the data from FEA simulations involved eliminating internal points. (Fig. 2a) Owing to the heterogeneity introduced by the tetrahedral mesh, the point cloud distribution lacked uniformity. We first employed a grid average method for downsampling, which ensured uniform point distribution. However, the resultant number of points varied, prompting us to further employ a random downsampling technique, adjusting all point cloud counts to "N" (the smallest count post-grid average downsampling within the dataset, ensuring minimal random deletions). (Fig. 2b) Subsequently, using the pre-deformation node positions, we centered and rotated the point cloud. This ensured that for 2D models, the point $(0, 0)$ was centered within a square aligned parallel to the x-y plane, and for 3D models, the point $(0, 0, 0)$ was central within a cube, with each face aligned parallel to the x-y, y-z, or x-z planes. Finally, we normalized the post-deformation point cloud data, constraining its range between -0.5 and 0.5. (Fig. 2c)

\begin{figure}
\centering
\includegraphics[width=14cm]{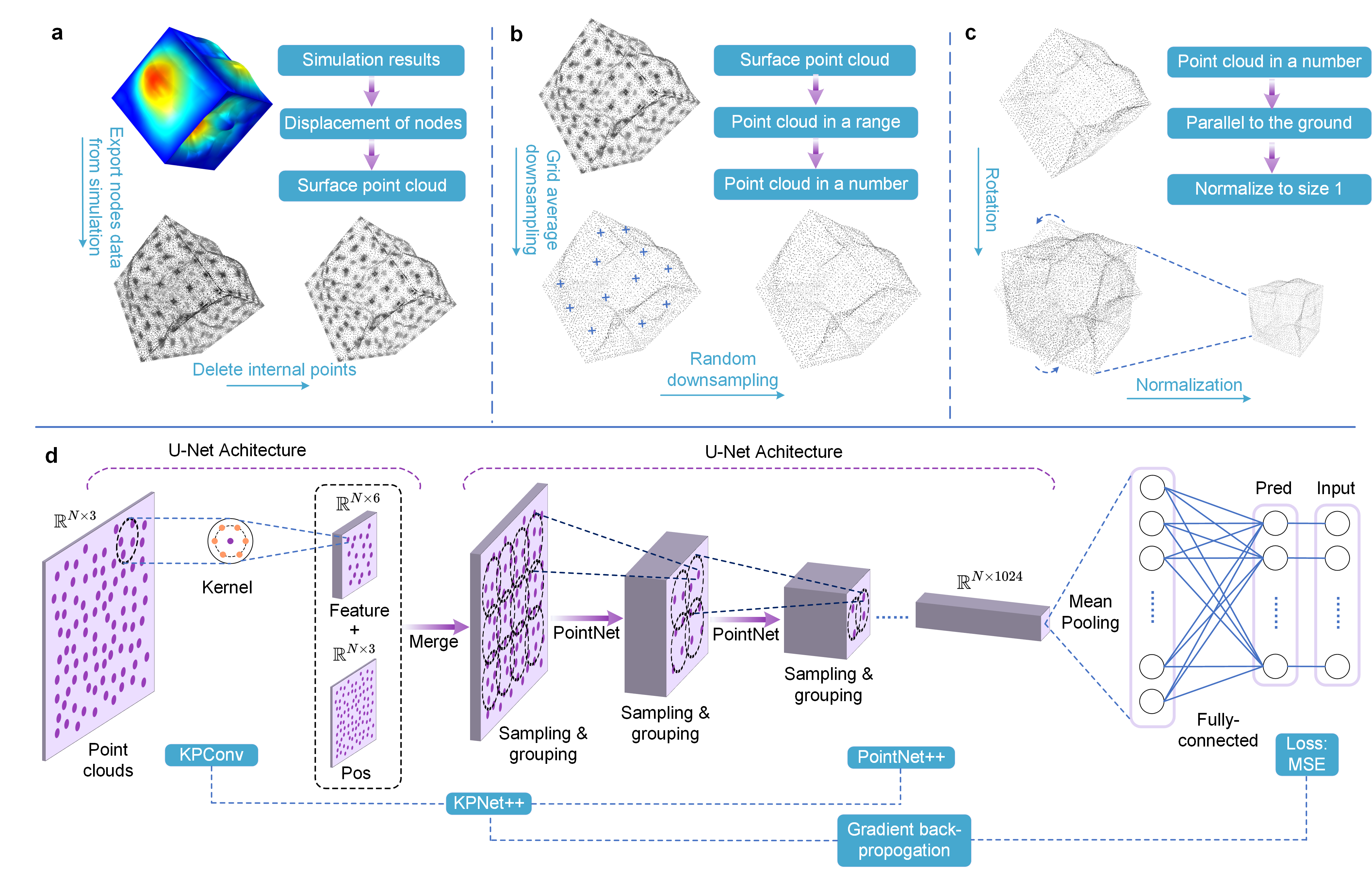}
\caption{\textbf{The framework of mapping point cloud from simulation results with the control inputs by using SMNet.} a, The procedures of extracting point cloud data from simulations. b, The downsampling strategy for point cloud data: including grid average downsampling to avoid the point concentration and random downsampling to ensure the number of points is the same. c, The point cloud rotation and normalization for training requirements. d, The architecture of SMNet for regression problems.}
\label{fig::figure2}
\end{figure}

\subsection{Architecture of SMNet}\label{sec3}

SMNet finds the regression between the control inputs of the structure and the resulting deformed geometry, which is represented as a point cloud. The first step of model training consists of processing the point cloud data $\mathcal{P} \in \mathbb{R}^{N \times 3}$ through a Kernel Point Convolution (KPConv) layer to identify the hidden features of the point cloud. (Fig. 2d) We construct a one-dimensional feature vector composed entirely of ones $\left\{\mathcal{F} \in \mathbb{R}^{N \times 1} \mid \mathcal{F_{\text{in}}}=\textbf{1} \right\}$, which serves as an initial feature for convolution with the point cloud. The whole KPConv part is built upon the foundational U-Net architecture, which comprises a sequence of encoding (down-sampling) layers followed by a symmetrical set of decoding (up-sampling) layers. Crucial to this structure are skip connections that directly link layers from the encoder to their counterpart layers in the decoder, ensuring the preservation and fusion of multi-scale features. Within its intermediate layers, unlike traditional convolutional operations that operate on standardized grids, KPConv is distinctively equipped with deformable kernels, allowing the convolutional kernels to adapt to more complex and varying geometric patterns, thereby enhancing the model's capacity to represent intricate spatial relationships in the data. This convolution process yields point cloud features as $\mathcal{F_{\text{out}}} \in \mathbb{R}^{N \times 6}$.

Subsequent to this, we combine the newly generated features $\mathcal{F_{\text{out}}} \in \mathbb{R}^{N \times 6}$ with the original point cloud $\mathcal{P} \in \mathbb{R}^{N \times 3}$ to a new dataset $\mathcal{P_\text{f}} \in \mathbb{R}^{N \times 9}$ and then input to the advanced PointNet++ architecture. The base architecture of PointNet++ is also built upon U-Net architecture similar to the KPConv. In each encoding and decoding layer, there is a sampling process by using spheres to reorganize the point set followed by a grouping process to integrate the centroid points of each sphere with the points in the neighborhood of centroid points. Subsequently, the aggregated data passes through a mini-PointNet network consisting of convolution, normalization, and ReLU activations for each layer. 

After progressing through the PointNet++ framework, the output feature $\mathcal{F_\text{o}} \in \mathbb{R}^{N \times 1024}$ is subjected to average pooling. Here, given that the point cloud data is extracted from simulation results, it is inherently devoid of noise. Moreover, the deformations present are rather continuous, lacking in pronounced local detail features. Consequently, we opt for average pooling over max pooling. Then, fully connected layers are used to connect the feature vector from pooling to the target 152-dimensional (216 in 3D ionic case) output vector with the ReLU as activation function of each layer. Since the model's predictions are continuous number from -1 to 1, the Mean Squared Error (MSE) is employed as the pivotal loss function. The gradients of this loss with respect to the model parameters are computed and used to update the model's weights. (The detailed description of SMNet is introduced in Methods)

\subsection{Model performance}\label{sec3}

\subsubsection{Ionic 2D low-profile PSM \& 3D PSM}\label{sec4}

For the 2D ionic planar shape memory (PSM) system, it is integrated with a $6\times 6$ array of square-shaped ionic actuators as shown in Fig. 3a. In experimental systems,\cite{wang2023passively} each of these actuators can be independently controlled. Intriguingly, these ionic actuators exhibit bidirectional deformation; they bend downward upon the application of a positive voltage to the upper electrode and vice versa. The data employed herein derives from the simulation results in Abaqus, as previously published by our group. In both simulations and physical experiments of the ionic 2D PSM, the central point of the square is held fixed, a strategy employed to maximize deformation amplitude. This fixation method, however, induces a pronounced x-y shift in the post-deformation point cloud, making the reliance solely on z-axis data inadequate for learning. For detailed simulation configurations and data extraction protocols, readers are directed to our earlier work.\cite{wang2023passively} Concerning this PSM, we have amassed a training set of 5000 samples and a test set comprising 100 samples. To shed light on model performance and error distribution, an error map of the 100 ground truth input vectors for PSM control with the model's predicted vectors was executed. These disparities were visually conveyed through a $6\times 6$ color map, where each section indicates the input error of the corresponding actuator on the actual PSM, as depicted in Fig 3b. In subsequent analyses, the predicted vectors were injected into the simulation model to recreate the PSM shapes. A comparative assessment between the point cloud data of these reproduced shapes and the test set unveiled the error cloud map, showcased in Fig. 3c. This error cloud map, with a resolution of $30\times 30$, aggregates and averages the error within each grid, color-coded to represent varying magnitudes of discrepancies. As evident from Fig. 3c, due to the centrality of the fixed point, discrepancies predominantly amass in the upper right and lower left quadrants, while the central region showcases minimal error.

It's worth mentioning that beyond the SMNet model proposed in this study, we have also adapted prevalent point cloud segmentation and classification architectures like 3DCNN\cite{ge20173d,li2018pointcnn}, PointNet\cite{qi2017pointnet}, PointNet++\cite{qi2017pointnet++}, KPConv\cite{thomas2019kpconv}, and RSConv\cite{liu2019relation} for our regression task, serving as benchmarks. Unlike these models, SMNet demonstrates a distinctive advantage in handling shape morphing 3D point cloud data. This data, derived from simulation nodes, consists solely of coordinate information and is stable with no noise. As this 3D data lacks additional features, SMNet initially employs KPConv to acquire local shape features of the point cloud beyond coordinates. Subsequently, it utilizes PointNet++ to integrate and learn from both coordinates and local shape features, making it particularly effective for this specific application. Upon completing training, the coefficient of determination, R2 score, gleaned from the test set, was selected as the performance metric. The bar figure of models' R2 scores is portrayed in Fig. 3d and the MLP model and SMNet model are highlighted. (Fig. 3d) In previous endeavors where merely z-displacement was employed—eschewing 3D point cloud data—the Multilayer Perceptron (MLP) achieved a modest accuracy of 0.8223\cite{wang2021design,wang2023passively}. A paradigm shift to utilizing point cloud data saw every model subjected to five independent training sessions. The resultant mean accuracies were as follows: 3DCNN at 0.8403, PointNet at 0.8655, KPConv at 0.9064, RSConv at 0.9319, and PointNet++ at 0.9336. Notably, our newly introduced SMNet culminated in an impressive accuracy of 0.9768—a surge of $15.45\%$ in the R2 score. (Fig. 3d) The details of MSE and Mean Absolute Error (MAE) are shown in Extended Data Table 2. We can find that the MSE of SMNet at 0.0078 mm is merely $13\%$ of the previous MLP's 0.0595 mm, indicating a significant decrease of more than 7.5 times.

\begin{figure}
\centering
\includegraphics[width=14cm]{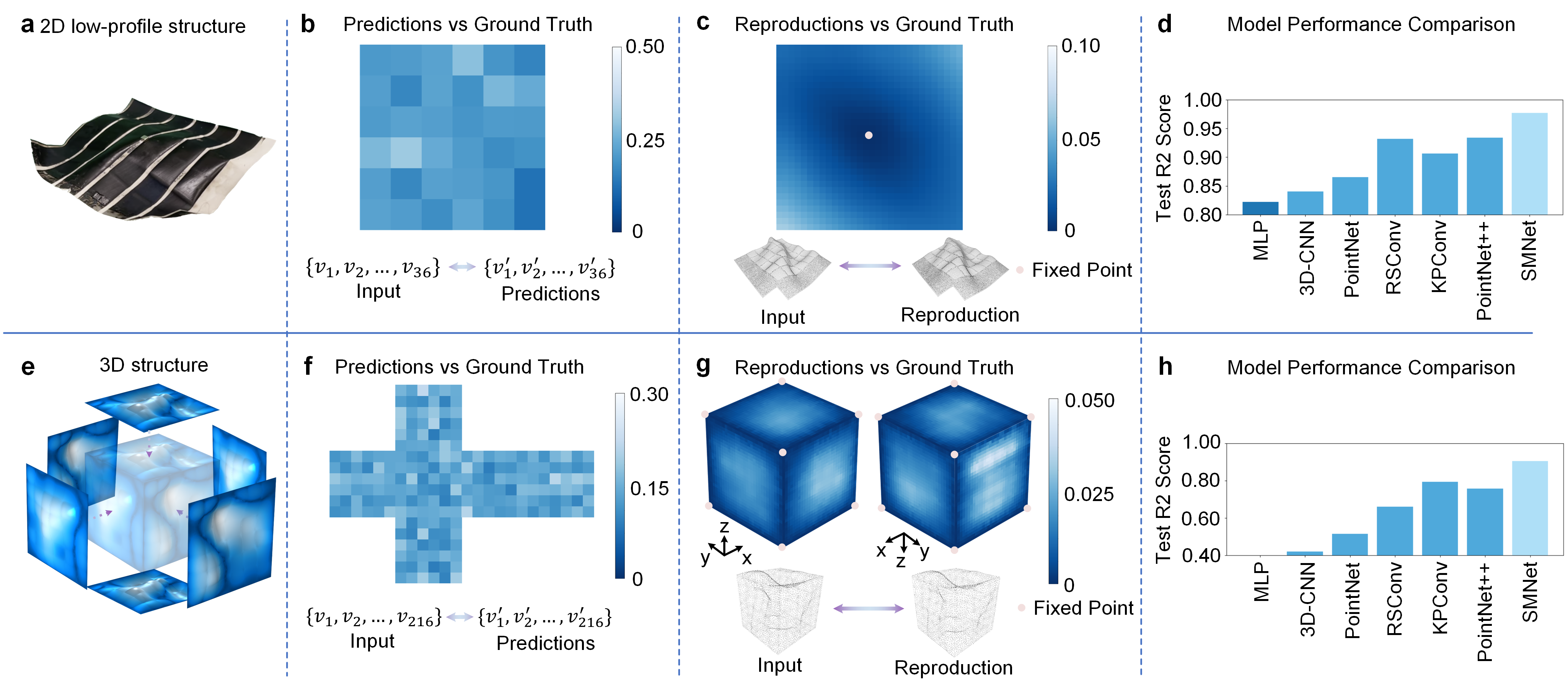}
\caption{\textbf{The model performance of ionic 2D low-profile PSM and 3D PSM.} a, The physical image of ionic 2D low-profile PSM proposed in\cite{wang2023passively}. b, The error map between model predictions and the ground truth control input vectors of ionic 2D low-profile PSM. c, The error map of ionic 2D low-profile PSM showcasing the point cloud of reproduced shapes with the ground truth point cloud. d, The comparison of R2 score across various training models for ionic 2D low-profile PSM. e, The exploded image of ionic 3D PSM assembled by 6 pieces of ionic 2D low-profile PSM. f, The unfolded error map between model predictions and the ground truth control input vectors of ionic 3D PSM. g, The 3D error map of ionic 2D low-profile PSM showcasing the point cloud of reproduced shapes with the ground truth point cloud. There are two angles of view to show the entire 6 surfaces of the cube. h, The comparison of R2 score across various training models for ionic 3D PSM.}
\label{fig::figure3}
\end{figure}

To validate the performance of SMNet for 3D PSM, we implemented a virtual model of a cube composed of six 2D low-profile PSMs, with each face featuring an independent $6\times 6$ array of ionic actuators. The simulation principle for this 3D PSM echoes that of the 2D counterpart. Through COMSOL, we conducted simulations with 20000 randomized control vectors and extracted the corresponding post-deformation point cloud data. Additionally, another 100 simulations were executed to serve as the test set. We compared the 100 input vectors from the test set with the model's predicted vectors. To offer an unambiguous view of the error for each actuator on every face, Fig. 3f unfolds the cube, with the topmost layer representing the cube's upper face, the four intermediary layers showcasing the lateral faces, and the bottommost layer representing the cube's base. Furthermore, the predicted control vectors were fed into the simulation model to recreate the PSM deformations. The disparity between the point cloud data of these reproduced shapes and the test set materialized as an error cloud map, as depicted in Fig. 3g. Given our data's simulation origin, we possessed point cloud coordinates both pre and post-deformation. Based on the pre-deformation cube coordinates, each face was segmented into $30\times 30$ smaller squares. The average error between the reproduced data and test set data within each square was computed and color-coded on the error cloud map. As illustrated in Fig. 3g, we presented the cubic error cloud map from two distinct angles, ensuring visibility of the discrepancies across all six faces. Given that the eight vertices of the ionic 3D PSM cube were held fixed during simulation, the peripheral errors were minimal, with the bulk of discrepancies centralized on the squares' central regions. The average deformation magnitude for each face was around 0.4, while the peak error value was 0.03. A comparative assessment with other models was also undertaken, with R2 scores illustrated in Fig. 3h. The details of MSE and MAE of each model are shown in Extended Data Table 2.

\subsubsection{Pneumatic 3D PSM \& thermal 3D PSM}\label{sec4}

In this study, we aim to propose a universally applicable, model-free technique for controlling all 3D shape morphing devices. To demonstrate this generality, we investigated two additional actuation mechanisms that are common in soft robotics: thermal actuation and pneumatic actuation.

Firstly, we investigated thermal actuator arrays whose deformation principle is based on volume change, with paraffin wax serving as the primary material due to its linear and significant thermal expansion and contraction within certain temperature bounds.\cite{mann2018modeling} In our demonstration, each thermal actuator was fashioned as a $1\times 1\times 1$ cube, with 152 of these units adorning the surface of a larger $6\times 6\times 6$ cube. Notably, the core of this assembly was a static $4\times 4\times 4$ passive cube unaffected by temperature-driven volume alterations. The design choice to place actuators solely on the surface stemmed from our focus on capturing the external point cloud transformations, as real-world data predominantly provides external surface point clouds. Internal actuator modifications were deemed to have limited and unclear impacts on this external point cloud. Each of the 152 surface actuators could be individually temperature-controlled, with thermal cross-talk negated by a thin insulative layer ensuring no inadvertent temperature-driven effects on adjacent units. (Fig. 4a) 

Secondly, we delved into pneumatic soft actuator arrays operating on a bulking deformation principle. Similarly, 152 $1\times 1\times 1$ pneumatic chambers were placed on the $6\times 6\times 6$ cm cube's surface, each capable of independent pressure modifications. Detailed simulation setups for both mechanisms will be elaborated upon in the Methods section. (Fig. 4b) 

For both mechanisms, we generated 20000 point clouds through FEA simulations to serve as our training dataset, while designating 100 sets as our test dataset. After getting the pre-trained model, we employed it to forecast our test dataset, resulting in 100 predictions. These predicted values were then integrated into their respective FEA models to generate point clouds of the reproduced shapes. A comparative analysis, akin to the methodology outlined in the previous section, was conducted between the predictions and test dataset, as well as input point clouds and reproduced point clouds. The findings from this comparative study are depicted in Fig. 4c-f. In the case of the thermal mechanism, the fixed point is situated at the model's center, allowing for an unhindered movement of all points on the surface. Consequently, the errors of predictions and the reproduced point clouds in relation to the ground truth exhibit a uniform distribution across the surface. (Fig. 4c,d) However, for the pneumatic mechanism, its fixed points align with those of the ionic mechanism, situated at the eight vertices of the cube. However, unlike the ionic mechanism where each actuator deforms relatively independently, the bulking deformation principle inherent to the pneumatic mechanism results in substantial coupling between the deformations of actuators. As a consequence, the deformation near the 12 edges is relatively limited, leading to a slight decline in prediction accuracy for the edge chambers compared to the central chambers. (Fig. 4e) Nevertheless, for the reproduced point clouds, given the negligible deformation along the edges, the primary deviations are predominantly centered but remain minimal. (Fig. 4f) Also, the overall deformation of the pneumatic mechanism possesses lower surface complexity compared with the other two mechanisms.

Given that KPConv and PointNet++ are sublayers of SMNet, we specifically compared the performance of these three models across different mechanisms. As shown in Fig. 4g, we laid out the error maps of the reproduced point cloud and the ground-truth point cloud into six faces, arranged in two rows. Additionally, we compared each dimension of the predicted input vector with the ground-truth input vector, and linearly displayed the error of each dimension below the point cloud error maps. Since all the figures are of the same scale, it is evident that SMNet demonstrates the best performance in terms of prediction accuracy. The specific training data (MSE, MAE, R2 score) for the three models across the three mechanisms is shown in Extended Data Table 3.

We also evaluated the number of trials required for training, as shown in Extended Data Fig. 1a. For 2D PSM devices, a dataset of 5000 trials proved sufficient for different models to achieve high prediction accuracy. The significant improvement in prediction accuracy was observed between 1000 to 3000 trials. For 3D PSM devices as depicted in Extended Data Fig. 1b, which entail higher complexity, larger training sets are necessary. In the case of the Thermal mechanism, where actuator coupling is relatively low, a dataset of 10000 trials suffices to meet training requirements. However, for the Ionic mechanism, an example of high actuator coupling requiring the most inputs, a training set of 20000 trials is needed to reach an acceptable accuracy level.

\begin{figure}
\centering
\includegraphics[width=14cm]{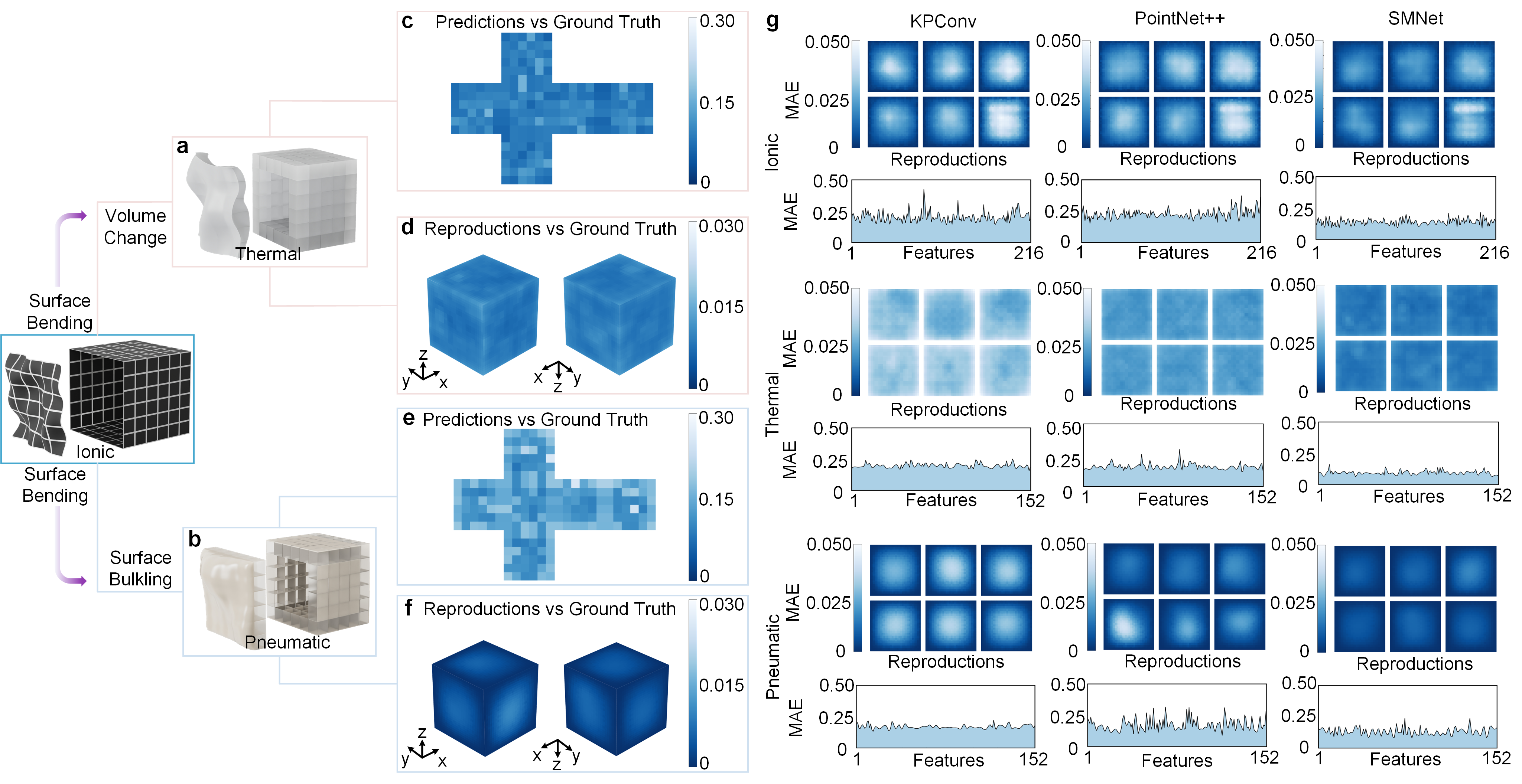}
\caption{The model performance of thermal 3D PSM and pneumatic 3D PSM and the comparison between the model performance of SMNet with KPConv and PointNet++. a, Expanding from ionic 3D PSM based on the bending principle to thermal 3D PSM based on volume change. b, Expanding from ionic 3D PSM based on the bending principle to pneumatic 3D PSM based on surface buckling. c \& e, the unfolded error map between model predictions and the ground truth control input vectors of thermal 3D PSM and pneumatic 3D PSM, respectively. d \& f, the 3D error map of thermal 3D PSM and pneumatic 3D PSM showcasing the point cloud of reproduced shapes with the ground truth point cloud. There are two angles of view to show the entire 6 surfaces of the cube. f, the model performance comparison between KPConv, PointNet++, and SMNet. We laid out the error maps of the reproduced point cloud and the ground-truth point cloud into six faces, arranged in two rows. Additionally, we compared each dimension of the predicted input vector with the ground-truth input vector, and linearly displayed the error of each dimension below the point cloud error maps.}
\label{fig::figure4}
\end{figure}

\subsection{Inverse demos for 3D PSM model-free control}\label{sec4}

In the context of shape-morphing devices, the paramount capability is inverse controllability. This implies that one can input a desired target shape, and the device can transform to match it. To validate the broad applicability of our SMNet for 3D shape morphing control, it is imperative for the system to adapt to any physical shape found in the real world. This aspiration aligns with one of the ultimate objectives in the domains of soft robotics, biomimetic robots, and haptic devices.

To facilitate this, we manually molded clay to create a target 3D shape. Using a 3D scanner, we captured the resultant shape of the clay in the form of point cloud data. After preprocessing the data as illustrated in Fig. 5b, it was input into our pre-trained model. The output prediction from this model is the control vector for the 3D shape morphing device. For instance, under the ionic mechanism, the output represents voltage values of each pixel (216-dimensional vector), for the thermal mechanism, it is the temperature values for each small cube (152-dimensional vector), and for the pneumatic mechanism, it's the air pressure values within each chamber (152-dimensional vector). These control vectors are then separately input into the FEA models of these mechanisms, revealing the morphed outcomes upon specifying a target shape.

Demos 1 and 2 present two shapes of varying complexities formed by manually molding the clay. The 3D scanned images of these physical shapes and the reproduced point cloud representations from the three mechanisms are showcased in Fig. 5a. To elucidate the deformation effect in the reproduced point cloud images, we calculated the displacement values of each point before and after deformation and represented this using a color gradient. Through the colored point cloud, it can be seen that all three mechanisms are capable of reproducing the main features of Demo 1 and 2.

However, since manually molded clay might lack intricate detail, in Demo 3, we employed Autodesk Maya software to create a 3D model, which prominently features the word "PURDUE" protruding on its six faces. The reproduced point cloud results of the three mechanisms for this design are depicted in the third column of Fig. 5a. It's evident from these reproduced point clouds that the ionic and thermal mechanisms, owing to the relative independence of each actuator and minimal mechanical coupling with adjacent actuators, offer superior programmability. The shapes from all three demos were impeccably replicated by these mechanisms. Conversely, the pneumatic mechanism, characterized by significant mechanical coupling between actuators, could only reproduce a relatively simple shape (Demo 1). It manifested discernible deviations from the target shape in Demo 2, and Demo 3 was entirely beyond its morphing capacity, resulting in a shape bearing scant resemblance to the original. The detailed procedures of demos are shown in Supplementary Video 1.

To quantitatively assess the fidelity of the three mechanisms in replicating the three demos, we subjected the original point clouds and their respective reproductions to a similarity analysis. In the realm of point cloud similarity metrics, three primary distances are prominently employed: Chamfer Distance (CD)\cite{wu2021balanced}, standard deviation of distance, and Hausdorff Distance (HD)\cite{memoli2004comparing}. The detailed explanation of these metrics is illustrated in Methods. The specific data for these three metrics are displayed in Fig. 5c. Due to the pneumatic mechanism's Demo 3 failing to achieve the target shape, its error significantly exceeds that of the other cases.

We also explored the feasibility of achieving real-time control with our developed models. To this end, we compared the training duration and demonstration execution times for three distinct models, as detailed in Extended Data Table 4. While the training time of these models necessitates a considerable time investment, executing pre-trained models requires just over one second. This response time falls below the actuation times of both ionic and thermal actuators. Since the control execution is faster than the physical actuation of the device,  the model can be used for real-time control.

Furthermore, our analysis revealed that predicting outcomes for 100 sets of data in parallel only incurs an additional delay of approximately 0.5 seconds compared to processing a single data set. This suggests that if a series of target shapes can be input simultaneously, the real-time responsiveness of the control system could be significantly enhanced. Therefore, the results lay a solid foundation for implementing real-time control in 3D shape morphing devices, thereby expanding their practical applications in various domains.

\begin{figure}
\centering
\includegraphics[width=14cm]{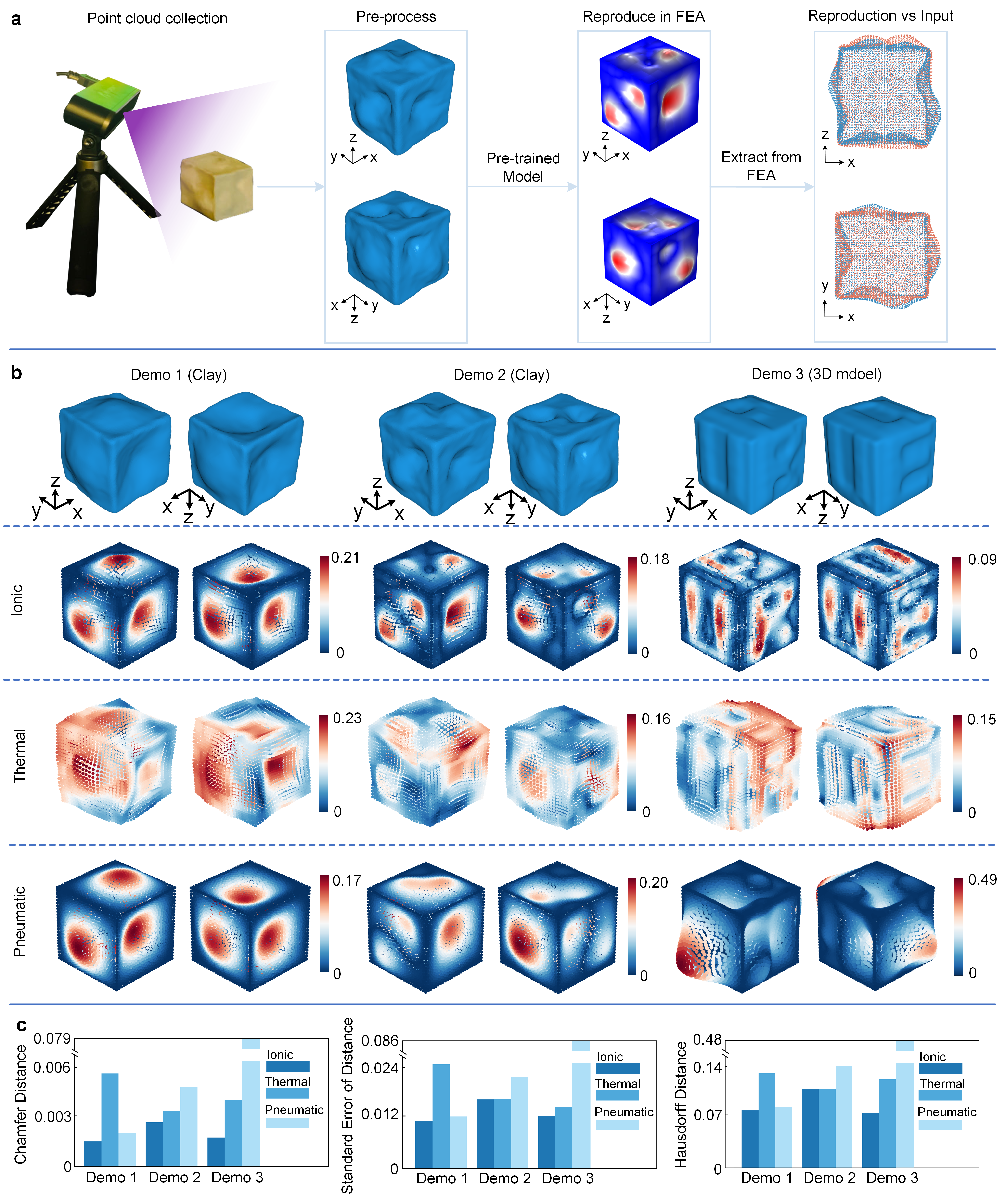}
\caption{\textbf{The demonstration of SMNet on inverse control of 3D shape morphing devices.} a, The detailed procedures of the demonstrations. b, The reproduced point cloud of 3 different mechanisms. `Demo 1' and `Demo 2' present two shapes formed by manually molding the clay. `Demo 3' is made by software with high surface complexity. c, The similarity between the reproduced point cloud with the target point cloud by using Chamfer Distance (CD), standard deviation of distance, and Hausdorff Distance (HD). All of the data has been normalized to 1. To facilitate a better comparison among the other cases, the bars for pneumatic actuators employ a truncated axis.}
\label{fig::figure5}
\end{figure}

\section{Discussion}\label{sec12}

This study introduces a universal approach to control 3D shape morphing devices across various actuation principles. By leveraging deep learning techniques and point cloud data, we have developed a point cloud regression model SMNet to map desired 3D shapes to the high-dimensional input vectors that represent control inputs, providing a model-free control methodology that can be applied to shape morphing devices with various actuation mechanisms. SMNet's ability to handle intricate geometric couplings and deformations enables precise control over shape morphing devices. SMNet has demonstrated superior performance compared to existing models in predicting control vectors for various 3D shape morphing devices. This capability enables inverse control, where physical objects and complex virtual 3D models are replicated with high fidelity using different actuation mechanisms.

In this manuscript, the training data was generated from FEA models because it offers an automated approach to generate large datasets that are free from noise and non-idealities such as manufacturing variations. Our approach is enabled by advancements in FEA that have significantly improved the accuracy of simulations.\cite{xavier2021finite,boyraz2018overview} However, our proposed method would also be compatible with source data from physical devices, with the drawback of long times required to collect the training data.

Compared to traditional mathematical modeling, our method offers significant advantages. Mathematical models become increasingly challenging as the coupling between actuators increases and as the shapes become more complex, such as moving from 2D to 3D shape morphing. The development of mathematical models is time-consuming and often requires numerous assumptions that can diminish the precision of inverse control. Our proposed methodology simplifies this process by requiring only the creation of a simulation model, which is then used to generate a dataset through repeated computations with different inputs. This dataset is subsequently trained using our SMNet. For instance, in the case of the ionic 3D PSM explored in this paper, the simulation time for a single result was approximately 5 minutes. We utilized a server powered by two AMD 7H12 CPUs for parallel computations, acquiring 20000 data sets within five days. The training phase, when conducted using multiple H100 parallel processors, could be condensed to approximately 3-4 days. As such, a fully operational control model for a highly complex 3D PSM device can be developed in around ten days, significantly streamlining the process compared to traditional methods.

In an inverse prediction task, the error includes a contribution from the accuracy of the prediction model as well as a contribution from the limitations of the actuator mechanism. For example, an actuator that can only achieve a small bending deformation will not be able to produce a shape that requires sharp features. The demos that were prepared with three different actuator mechanisms (ionic, thermal, and pneumatic) therefore have different abilities to reproduce the target shapes based on the limitations of their actuation mechanisms. The complexity of a surface can be quantified by the variance in normal vectors, as described in the methods. The surface complexity of the demos and the reproduced shapes are included in Extended Data Table 5. The pneumatic actuation mechanism consistently achieves a surface complexity that is significantly lower than the target shape. This emphasizes a key advantage of this generalized approach for shape morphing control: it allows comparison of different actuation mechanisms to facilitate the selection of an appropriate actuation mechanism for the target geometry.

In this work, we use SMNet to find the relationship between the input control parameters of a 3D actuator array and the point cloud describing the deformed geometry. Modified versions of SMNet may find widespread value in predicting the deformation of other soft continuum structures, such as the deformation of biological systems during growth based on input parameters such as the location and type of cells. Future work can therefore investigate the use of SMNet with cloud point datasets that include points within the interior of a structure that can be extracted from 3D imaging techniques.

%The results from the demo show the control method is effective within the programmability limits of the 3D PSM devices. Therefore, how to quantitatively define and evaluate the programmability of a 3D PSM device could be a potential research opportunity. It should depend on the working range of each actuator and the coupling of actuator arrays. One potential way to evaluate the programmability is to calculate the complexity of different shapes that a shape morphing device can morph into. The details are shown in Methods.

In summary, this research paves the way for advanced development in soft robotics, particularly in areas requiring intricate 3D shape transformations. The universality and effectiveness of our proposed control method hold great promise for 3D PSM devices in human-machine interfaces, haptic technology, wearable gadgets, and beyond. Still, for realizing a fully programmable, controllable, and capable of real-time control 3D PSM device, further research in actuator materials, fabrication techniques, and addressing methods are crucial to bridge the gap between theoretical control and practical application of 3D PSM devices.

\section{Methods}\label{sec11}

\subsection{FEA simulation of actuation mechanisms}

The simulation of ionic actuators is described in detail in our previous work.\cite{wang2023passively} The architecture of the ionic actuator adopts a sandwich configuration. The outer layers are composed of a conductive electrode that swells in response to a voltage, for which we use the materials properties of polypyrrole (PPy). The middle layer is an ionic conductor and electrical insulator, for which we use the materials properties of porous poly(vinylidene fluoride) (PVDF). In our simulation, the thickness of the central PVDF layer is set at 110 microns, and that of the electrode is 20 microns. Material-wise, the Young's modulus for PVDF is 2.45 GPa and for PPy, it's 2 GPa. A Poisson Ratio of 0.25 was uniformly assigned to both materials. For the sake of simulation in ionic actuators, thermal expansion is conventionally used as a surrogate for its electrical expansion. Drawing from our prior experimental data,\cite{wang2023passively} the thermal expansion coefficient for PPy is defined as 0.05, while for PVDF it's established at $1.2e-6$ based on its inherent properties. Both materials have their specific heat capacities set at $4200 Jkg^{-1}K^{-1}$. Structurally, the cubic framework is assembled from six individual square panels, each hosting a $6\times 6$ array of discrete ionic actuators. Every standalone ionic actuator is square-shaped, with a side-to-gap ratio of 10:1. This simulation was conducted in COMSOL, opting for a Tetrahedral mesh with the default coarser size setting. The applied voltage on a pixel was assigned as a boundary condition with half of the requisite voltage applied to the outer PPy regions and half to the inner sections. For instance, if a pixel's target voltage is -0.6 V, the distribution would be -0.3 V at the outer surface and 0.3 V at the inner surface. Mechanical boundary conditions consisted of fixed constraints at the eight vertices of the cube.

As to the thermal actuator array, we chose Abaqus as the simulation software. In this case, we aim to simulate the shape morphing caused by temperature-induced volume changes. The structure consists of a cubic geometry with each face consisting of $6\times 6$ addressable pixels. Given our focus on surface deformations, only the pixels on the cube's surface are actuated, while the interior of the cube is made of an undeformable and temperature-insensitive material. For the thermal actuators on the surface layer, we employed materials exhibiting linear thermal expansion properties, exemplified by paraffin wax actuators. The thermal expansion coefficient of the actuating material was chosen as 0.4 to normalize the deformation range to input temperatures in the bound of $\pm$ 1 degree Celsius, and we assume it has high thermal conductivity that allows the temperature in this cube to be uniform. To ensure that the temperature of each actuating pixel is independent, we added a very thin layer of material with super-low thermal conductivity between the pixels. Those intersecting sheets have a thickness of $0.5\%$ of the thermal actuator length. To simulate the thermal expansion of this actuator array with unique temperatures assigned to each actuating pixel, a "Coupled Temperature-Displacement" step was created in Abaqus to analyze the Steady State response, with automatic incrementation, maximum number of increments set to 100, and a minimum increment size of $1e-5$, with other setting being defaulted. The unique temperatures informing the thermally isolated expansion of each actuator was addressed by iterating over each actuator's surfaces normal to the Z direction in the script and applying the chosen randomized temperatures as "Temperature Boundary Conditions" in Abaqus. As for mechanical boundary conditions, we apply an "Encastre Boundary Condition" at the very central non-actuating point of the $6\times 6\times 6$ grid, thus constraining the location without affecting the deformation due to the outer actuators. A very coarse mesh size of a quarter of the actuator length was chosen to reduce the time of a single simulation since the training requires 20000 trials, and the element type is chosen as "C3D8T". 

Within the pneumatic actuator array, we implemented a hollow structure in the core. Only on the six surfaces of the cube are $6\times 6$ chambers established. Owing to the shared chambers at the edges and corners, the total number of independently controlled chambers is 152. We used the materials properties of Sylgard 184, a common silicone elastomer used in soft robotics. Sylgard 184 has a typical Young's modulus of 2 MPa and a Poisson's ratio of 0.48. As this simulation was also executed in COMSOL, the mesh choice remained a Tetrahedral mesh with the default coarser size. The pneumatic inputs were normalized to a range between -1 and 1 to prevent gradient explosions during subsequent machine learning training phases. Additionally, constraints were affixed at the eight vertices of the cube.

\subsection{Principle of SMNet}

In the KPConv layer, we have the point coordinates $\{\mathbf{x}_i\}$ drawn from a point cloud $\mathcal{P} \in \mathbb{R}^{N \times 3}$. Correspondingly, we construct a feature vector $f_i$ for each point, initially filled with ones, denoted by the set $\{\mathcal{F} \in \mathbb{R}^{N \times 1} \mid f_i=\mathbf{1}\}$. We introduce a kernel within a predefined radius $r \in \mathbb{R}$, and define the neighborhood $\mathcal{N}$ of a point $x$ within this radius as $\mathcal{N}=\{\mathbf{x}_i \in \mathcal{P} \mid \|\mathbf{x}_i - \mathbf{x}\| \leq r\}$, where $\mathbf{x}_i$ and $\mathbf{x}$ denote the coordinates used to calculate Euclidean distances. The convolution of feature $f$ by the kernel $g$ centered at $x$ is thus formulated as:
\begin{equation}
(\mathcal{F} \ast g)(x) = \sum_{x_i \in \mathcal{N}} g(x_i - x) f_i
\label{eq:convolution}
\end{equation}
The kernel function $g$ operates on the relative positions of neighboring points, which are computed as $\mathbf{y}_i = \mathbf{x}_i - \mathbf{x}$. Its domain is the sphere $\mathcal{B}_r^3 = \{\mathbf{y}_i \in \mathbb{R}^3 \mid \|d_i\| \leq r\}$. We represent kernel points as $\{\widetilde{\mathbf{x}}_k\}$ with the constraint $\{\widetilde{\mathbf{x}}_k \mid k < K\} \subset \mathcal{B}_r^3$, where $K$ signifies the total number of kernel points. The associated weight matrices that project features from the input dimension, which is 4 (considering a supplementary 1D feature vector of ones appended to the point coordinates), to an output dimension $D_{\text{out}}$ are expressed as $\{W_k \mid k < K\} \subset \mathbb{R}^{4 \times D_{\text{out}}}$. Accordingly, the kernel function $g$ can be formulated as \cite{thomas2019kpconv}:
\begin{equation}
g(\mathbf{y}_i) = \sum_{k < K} h(\mathbf{y}_i, \widetilde{\mathbf{x}}_k) W_k
\label{eq:kernel_function}
\end{equation}
Here, $h$ signifies the correlation between kernel points $\widetilde{x}_k$ and the relative position $y_i$. A linear correlation function is applied as:
\begin{equation}
h(\mathbf{y}_i, \widetilde{\mathbf{x}}_k) = \max\left(0, 1 - \frac{\|\mathbf{y}_i - \widetilde{\mathbf{x}}_k\|}{\sigma}\right)
\end{equation}
where $\sigma$ denotes the influence distance of the kernel points. In this study, $K$ is selected to be 15, and $\sigma$ is chosen as 1.5.

As mentioned above, the base architecture of KPConv is U-Net, which contains 3 encoding layers and 2 decoding layers, and there is a KPConv in each layer. In each encoding layer, multiscale precomputed data would go through the down convolution while in each decoding layer, upsample precomputed data would take the up convolution. The final output of KPConv layer is a 6D feature $\mathcal{F_{\text{out}}} \in \mathbb{R}^{N \times 6}$

In PointNet++ layer, the input is a pointset $\mathcal{P_\text{f}} \in \mathbb{R}^{N \times 9}$ combined the newly generated features $\mathcal{F_{\text{out}}} \in \mathbb{R}^{N \times 6}$ with the original point cloud $\mathcal{P} \in \mathbb{R}^{N \times 3}$ to a new point sets $\mathcal{P_\text{f}} \in \mathbb{R}^{N \times 9}$. As aforementioned, the PointNet++ layer contains the sampling, grouping, and PointNet process. In sampling process, the input point sets is $\mathcal{P_\text{f}} \in \mathbb{R}^{N \times 9}$. By utilizing the iterative farthest point sampling (FPS), a subset of points with number of $N_s$, $\mathcal{P_\text{n}} = \{\widetilde{\mathbf{x}}_n \mid n < N_s\} \subset \mathcal{P_\text{f}}$, can be chosen. In this subset, the point $\mathbf{x}_{ni}$ is the most distant point (in metric
distance) from the set $\mathcal{P_{\text{nj}}} = \{\widetilde{x}_{nj} \mid j < i\} \subset \mathcal{P_\text{n}}$ with regard to the rest points\cite{eldar1997farthest}. The pseudocode is shown below:

\begin{algorithm}
\caption{Farthest Point Sampling (FPS)}
\begin{algorithmic}[1]
\Procedure{FPS}{$\mathcal{P_\text{f}}$, $N_s$}
    \State $\mathcal{P_\text{n}} \gets \emptyset$
    \State Select a random point $x_i$ from $\mathcal{P_\text{f}}$ and add it to $P_\text{n}$
    \ForAll{$x \in P_\text{f}$}
        \State $[d] \gets \|x_i - x\| $
    \EndFor
    \While{$|\mathcal{P_\text{n}}| < N$}
        \State $x_{farthest} \gets \text{arg max}([d]) (x \in \mathcal{P_\text{f}} \setminus \mathcal{P_\text{n}})$
        \State Add $x_{farthest}$ to $\mathcal{P_\text{n}}$
        \ForAll{$x \in \mathcal{P_\text{f}} \setminus \mathcal{P_\text{n}}$}
            \State $[d] \gets \min([d], \|x_farthest - x\|)$
        \EndFor
    \EndWhile
    \State \Return $\mathcal{P_\text{n}}$
\EndProcedure
\end{algorithmic}
\end{algorithm}

As to grouping, we group the original point set $\mathcal{P_\text{f}} \in \mathbb{R}^{N \times 9}$ and the coordinates the centroids of the subset $\mathcal{C} \in \mathbb{R}^{N_s \times 3}$. The output of grouping is could be $\mathcal{G} \in \mathbb{R}^{N' \times K \times 9}$, where K is the is the number of points in the neighborhood of centroid points.\cite{qi2017pointnet++}

In the PointNet layer, the coordinates of points in a local region are firstly translated into $\mathbf{y}_i = \mathbf{x}_i-\mathbf{x}$ where $\mathbf{x}$ is the coordinates of the central points. 

\begin{equation}
f\left((\mathbf{y}_1, \mathbf{y}_2, \ldots, \mathbf{y}_n\right)=\operatorname{MLP}(\max(\operatorname{MLP}(\mathbf{y}_i)))
\end{equation}

where $\operatorname{MLP}$ refers to multi-layer perception (MLP) networks. 

After going through multiple aforementioned steps, the final output is the feature vector $\mathcal{F_{\text{out}}'} = (\mathbf{f}_1, \mathbf{f}_2, \ldots, \mathbf{f}_n)$ with dimensional of 1024. ($\mathcal{F_{\text{out}}'} \in \mathbb{R}^{N \times 1024}$) We utilize average pooling for these output feature vectors.

\begin{equation}
\overline{\mathbf{f}_i}=\frac{1}{\left|N\left(p_i\right)\right|} \sum_{j \in N\left(\mathbf{x}_i\right)} \mathbf{f}_j
\end{equation}
where $N\left(\mathbf{x}_i\right)$ is the number of neighborhood points of point $\mathbf{x}_i$. 

The final step of SMNet is the fully-connected layer to map the abstracted features of point cloud to the ground truth features. The loss function of the regression procedure is chosen as MSE:

\begin{equation}
L(\mathbf{v}) = \frac{1}{N} \sum_{i=1}^N\left(\mathbf{v}_i-\hat{\mathbf{v}}_i\right)^2
\end{equation}
where $\mathbf{v}_i$ is output features from SMNet and $\hat{\mathbf{v}}_i$ is the ground truth features.

\subsection{Training setting-ups}

In our study, the training architecture was constructed using PyTorch. The model was trained on an RTX 3090 GPU. Due to memory constraints associated with the RTX 3090, we set the batch size to 8. For ionic 2D low-profile PSM, the epochs of training are 200 while for all 3D PSM cases, the epochs of training are 600. The optimization was performed using Stochastic Gradient Descent (SGD) with a learning rate of 0.1 and a momentum of 0.9.

\subsection{The evaluation methods of the similarity of point clouds}

CD is computed as the sum of squares of the average distances from point cloud $\mathcal{P_{\text{a}}} \in \mathbb{R}^{N \times 3}$ to $\mathcal{P_{\text{b}}} \in \mathbb{R}^{N \times 3}$ and vice versa. It provides a symmetrical similarity measure, aptly suited for scenarios like ours where point cloud orientation is ambiguous and bidirectional errors are of concern. The magnitude of CD is a direct reflection of the similarity between two point clouds. 

\begin{equation}
d_{\mathrm{CD}}(\mathcal{P_{\text{a}}}, \mathcal{P_{\text{b}}})=\frac{1}{|\mathcal{P_{\text{a}}}|} \sum_{a \in \mathcal{P_{\text{a}}}} \min _{b \in \mathcal{P_{\text{b}}}}\|a-b\|^2+\frac{1}{|\mathcal{P_{\text{b}}}|} \sum_{b \in \mathcal{P_{\text{b}}}} \min _{a \in \mathcal{P_{\text{a}}}}\|a-b\|^2
\end{equation}

The standard deviation of distance metric calculates the shortest distance from each point in point cloud $\mathcal{P_{\text{a}}}$ to its closest counterpart in point cloud $\mathcal{P_{\text{b}}}$, and then derives their standard deviation. This metric gauges the spread or variability of these distances. A minimal standard deviation suggests that the majority of points from one cloud to the other approximate the average distance, indicating that the disparity between the two point clouds is uniform. Conversely, a pronounced standard deviation intimates that certain points in one cloud have distances significantly exceeding the average distance to the other cloud, whereas distances for other points are substantially below the average. This could be indicative of regions where the point clouds align exceptionally well and other regions where they do not.

\begin{equation}
\sigma(\mathcal{P_{\text{a}}}, \mathcal{P_{\text{b}}})=\sqrt{\frac{1}{n} \sum_{i=1}^n\left(d_i-\mu(\mathcal{P_{\text{a}}}, \mathcal{P_{\text{b}}})\right)^2}
\end{equation}

HD identifies the maximum of the shortest distances from each point in point cloud $\mathcal{P_{\text{a}}}$ to point cloud$ \mathcal{P_{\text{a}}}$. This process is reciprocated from $\mathcal{P_{\text{b}}}$ to $\mathcal{P_{\text{a}}}$, and the more considerable of the two values is established as the Hausdorff Distance. Serving as a symmetrical distance measure, HD accentuates the maximal discrepancy between two point clouds. Through these metrics, our objective was to rigorously discern the congruence between the original and reproduced point clouds, offering insights into the accuracy of the respective mechanisms.

\begin{equation}
d_\mathrm{HD}(\mathcal{P_{\text{a}}}, \mathcal{P_{\text{b}}})=\max \left(\max _{a \in \mathcal{P_{\text{a}}}} \min _{b \in \mathcal{P_{\text{b}}}}\|a-b\|, \max _{b \in \mathcal{P_{\text{b}}}} \min _{a \in \mathcal{P_{\text{a}}}}\|a-b\|\right)
\end{equation}

\subsection{Complexity of different shapes}

For quantifying the complexity of different shapes, one potential way is to calculate the "variance of the normal vectors" on its surface. 
The computation process unfolds as follows:
First, a mesh of size $n\times n$ is created over the surface, where the mesh's fineness is determined based on the required detail level. At every mesh vertex, a normalized three-dimensional normal vector $\vec{n} = (n_x, n_y, n_z)$ is assigned, obtained by computing the gradient at each surface point. Then, for each of these normal vectors, the angle $\theta$ they form with their immediate x and y direction neighbors in the grid is calculated (refer to Extended Data Fig. 2). Following this, the variance of these angles is calculated, serving as an indicator of the surface's angle variation dispersion or complexity. Lower variance indicates a smoother or more uniform surface, while higher variance points to a more intricate or detailed surface.
Based on that, we could derive this variance from the training dataset and get the maximum value as the largest programmability of the device. The input target shape should possess a lower "variance of the normal vectors", otherwise, the device may not be able to reproduce it. 

In computing the shape complexity of target shapes and reproduced shapes shown in Fig. 5, we re-grid the point cloud data and calculate the surface complexity of the six faces of the cube separately. Then, we take their average value as the final complexity of that shape.

\section{Extended Data}

\begin{itemize}
    \item Extended Data Fig. 1: The relationship between the number of trials for training and the R2 score.
    \item Extended Data Fig. 2: A potential way to quantify programmability of a PSM device.
    \item Extended Data Table 1: The MSE, MAE and R2 score of different models for ionic 2D PSM devices.
    \item Extended Data Table 2: The MSE, MAE and R2 score of different models for ionic 3D PSM devices.
    \item Extended Data Table 3: The training time and executing time of KPConv, PointNet++ and SMNet for all three kinds of 3D PSM devices.
    \item Extended Data Table 4: The training time and executing time of KPConv, PointNet++ and SMNet for all three kinds of 3D PSM devices.
    \item Extended Data Table 5: The complexity of the target shapes and reproduced shapes.
\end{itemize}

\begin{figure}
\centering
\includegraphics[width=14cm]{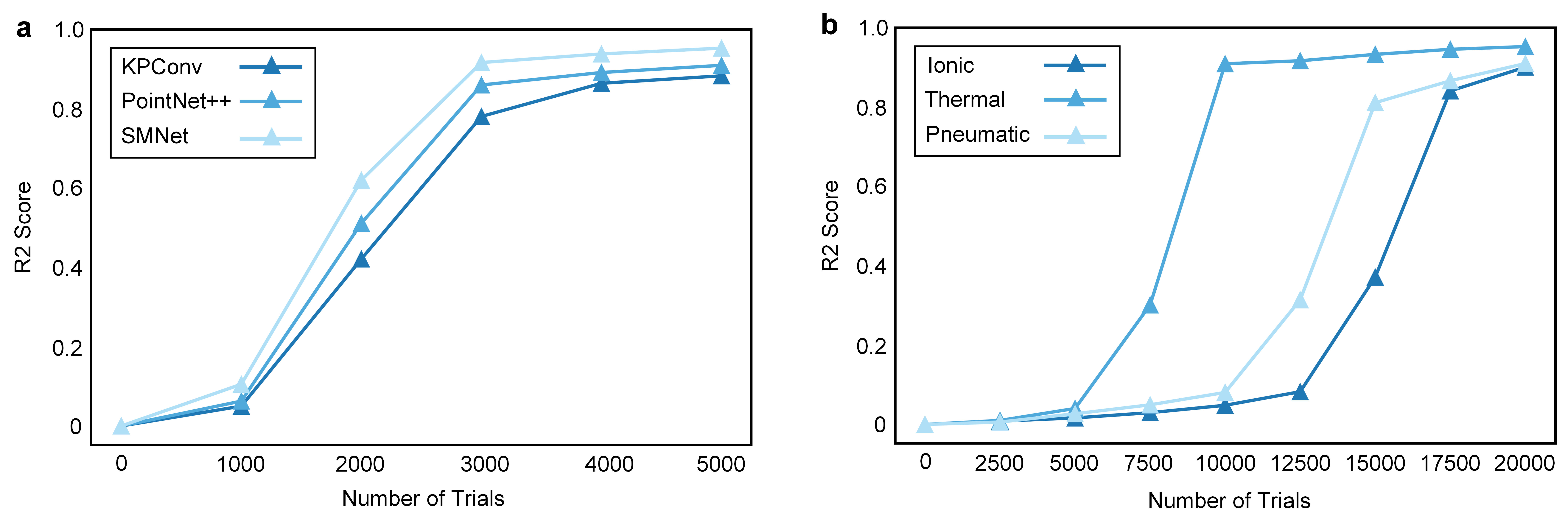}
\captionsetup{labelformat=empty}  
\caption{\textbf{Extended Data Fig. 1: The relationship between the number of trials for training and the R2 score.}  a, The R2 score vs the number of trials for ionic 2D PSM device under 3 different kinds of models. b, The R2 score vs the number of trials for all kinds of 3D PSM devices under the proposed SMNet.}
\label{fig::extended1}
\end{figure}

\begin{figure}
\centering
\includegraphics[width=14cm]{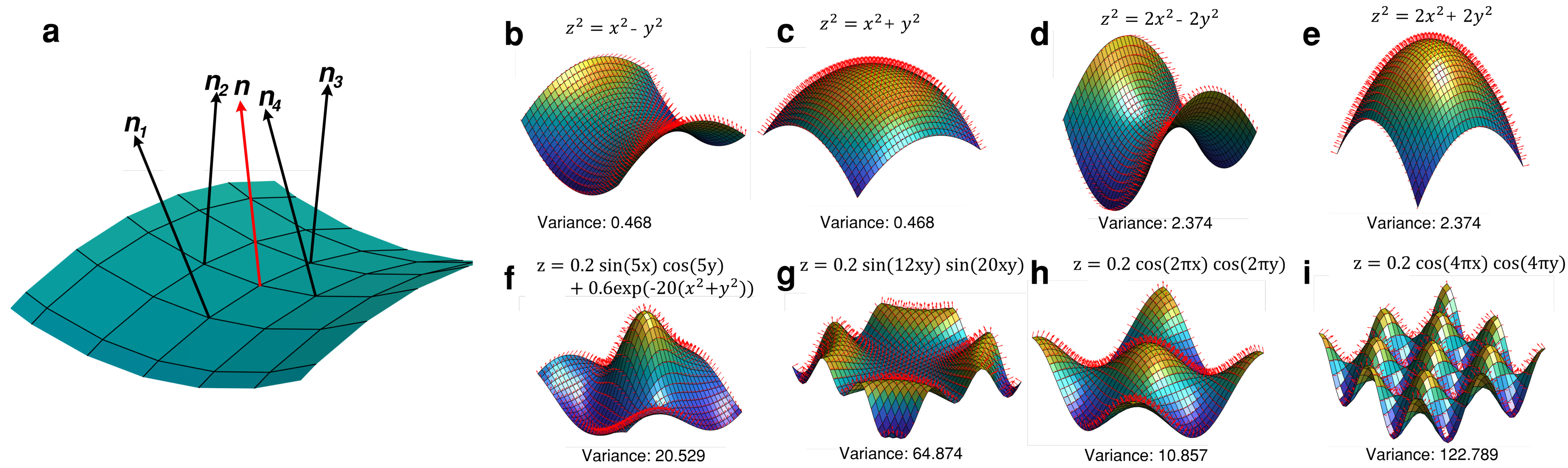}
\captionsetup{labelformat=empty}
\caption{\textbf{Extended Data Fig. 2: A potential way to quantify programmability of a PSM device.}  a, A potential way to define the complexity of a surface for by calculating the "variance of the normal vectors" on a surface. b, The standard saddle surface with variance of 0.468. c, The standard dome surface with variance of 0.468. d, A saddle surface with larger magnitude that has variance of 2.374. e, A dome surface with larger magnitude that has variance of 2.374. f \& g, Surfaces generated by random equations with variance of 20.529 and 64.874, respectively. h, A wavy surface with fewer peaks and valleys that has variance of 10.857. i, A wavy surface with more peaks and valleys that has variance of 122.789. The saddle and dome surfaces with the same magnitude have the same shape complexity, which makes sense since their formula only differs in the addition and subtraction signs. For wavy surfaces, the one with more peaks and valleys obviously has more ‘details’ on its surface, so its shape complexity is significantly larger than the other one.}
\label{fig::extended1}
\end{figure}

\begin{table*}[h]
\centering
\captionsetup{labelformat=empty}
\caption{\textbf{Extended Data Table 1: The MSE, MAE and R2 score of different models for ionic 2D PSM devices.} }
\label{table:model_performance}
\begin{tabular}{lccc}
\hline
Model & MSE (mean) & MAE (mean) & R2 score (mean) \\
\hline
MLP & 0.0595 & 0.1818 & 0.8223 \\
Linear & 0.1935 & 0.3545 & 0.4267 \\
3D-CNN & 0.0541 & 0.1773 & 0.8403 \\
PointNet & 0.0455 & 0.1583 & 0.8655 \\
RSConv & 0.0229 & 0.1183 & 0.9319 \\
KPConv & 0.0315 & 0.1401 & 0.9064 \\
PointNet++ & 0.0224 & 0.116 & 0.9336 \\
SMNet & 0.0078 & 0.0697 & 0.9768 \\
\hline
\end{tabular}

\end{table*}

\begin{table*}[h]
\centering
\captionsetup{labelformat=empty}
\caption{\textbf{Extended Data Table 2: The MSE, MAE and R2 score of different models for ionic 3D PSM devices.} }
\label{table:updated_model_performance}
\begin{tabular}{lccc}
\hline
Model & MSE (mean) & MAE (mean) & R2 score (mean) \\
\hline
MLP & N/A & N/A & N/A \\
Linear & N/A & N/A & N/A \\
3D-CNN & 0.1939 & 0.3546 & 0.4188 \\
PointNet & 0.1589 & 0.3163 & 0.5158 \\
RSConv & 0.1105 & 0.2639 & 0.6634 \\
KPConv & 0.0646 & 0.2000 & 0.8023 \\
PointNet++ & 0.0818 & 0.2241 & 0.7514 \\
SMNet & 0.0315 & 0.1398 & 0.9041 \\
\hline
\end{tabular}
\captionsetup{labelformat=empty}

\end{table*}

\begin{table*}[h]
\centering
\captionsetup{labelformat=empty}
\caption{\textbf{Extended Data Table 3: The training time and executing time of KPConv, PointNet++ and SMNet for all three kinds of 3D PSM devices.} }
\label{table:transposed_model_performance}
\begin{minipage}{\textwidth}
\centering
\caption*{Ionic 3D PSM}
\label{subtable:ionic}
\begin{tabular}{llll}
\toprule
Model     &        MSE (mean) &  MAE (mean) &  R2 score (mean) \\
\midrule
KPConv     &        0.0655 &      0.2011 &           0.7998 \\
PointNet++ &       0.0818 &      0.2241 &           0.7514 \\
SMNet      &          0.0315 &      0.1398 &           0.9041 \\
\bottomrule
\end{tabular}
\end{minipage}

\vspace{1em} % Add some space between the tables

\begin{minipage}{\textwidth}
\centering
\caption*{Thermal 3D PSM}
\label{subtable:thermal}
\begin{tabular}{llll}
\toprule
Model     &        MSE (mean) &  MAE (mean) &  R2 score (mean) \\
\midrule
KPConv     &         0.0575 &      0.1927 &           0.8226 \\
PointNet++ &      0.0409 &      0.1593 &           0.8734 \\
SMNet      &          0.0141 &      0.0928 &           0.9561 \\
\bottomrule
\end{tabular}
\end{minipage}

\vspace{1em} % Add some space between the tables

\begin{minipage}{\textwidth}
\centering
\caption*{Pneumatic 3D PSM}
\label{subtable:pneumatic}
\begin{tabular}{llll}
\toprule
Model &         MSE (mean) &  MAE (mean) &  R2 score (mean) \\
\midrule
KPConv     &         0.0449 &      0.1673 &            0.864 \\
PointNet++ &     0.0478 &      0.1701 &           0.8553 \\
SMNet      &          0.0286 &      0.1306 &           0.9132 \\
\bottomrule
\end{tabular}
\end{minipage}

\end{table*}

\begin{table*}[h]
\centering
\captionsetup{labelformat=empty}
\caption{\textbf{Extended Data Table 4: The training time and executing time of KPConv, PointNet++ and SMNet for all three kinds of 3D PSM devices.}}
\label{table:4}
\begin{tabular}{p{2cm}cccccc}
\hline
& \multicolumn{3}{c}{Training Time (s)*}  & \multicolumn{3}{c}{Executing Time (s)} \\
\hline
Model & Ionic & Thermal & Pneumatic & Ionic & Thermal & Pneumatic \\
\hline
KPConv & 1760.66 & 1507.15 & 1511.79 & 1.24 & 1.21 & 1.22 \\
PointNet++ & 337.88 & 301.28 & 302.44 & 0.99 & 0.92 & 0.93 \\
SMNet & 2053.36 & 1801.73 & 1803.01 & 1.45 & 1.42 & 1.42 \\
\hline
\end{tabular}
\\[10pt] % Adds some space after the table
\footnotesize{* 'Training Time' refers to the time required to run one epoch.}
\end{table*}

\begin{table*}[h]
\centering
\captionsetup{labelformat=empty}
\caption{\textbf{Extended Data Table 5: The complexity of the target shapes and reproduced shapes.} }
\label{table:4}
\begin{tabular}{cccc}
\hline
& Demo 1 & Demo 2 & Demo 3\\
\hline
Target Shape & 0.81 & 4.36 & 13.64\\

Reproduced Shape (Ionic) & 0.73 & 4.32 & 14.55\\

Reproduced Shape (Thermal) & 0.84 & 4.53 & 14.24\\

Reproduced Shape (Pneumatic) & 0.68 & 3.68 & 8.12\\
\hline
\end{tabular}

\end{table*}

\section{Acknowledgments}

This work was supported by the Purdue startup funding to Alex Chortos and by NSF award 2301509. There is no conflict of interest. Author contributions: Jue Wang proposed the idea of this paper. Jue Wang designed the algorithm, collected the data and write the manuscript. Alex Chortos directed the design process and revise the manuscript. Dhirodaatto Sarkar  contributed to the data collection and training process. Jiaqi Suo contributed to demonstrations.

\section{Data Availability}

The training data utilized in this study are openly available in the Zenodo repository at https://zenodo.org/records/10558468. Furthermore, the code supporting the findings of this research can be found at the following GitHub repository: https://github.com/wang5056/SMNet/tree/main.

\section{Supplementary Information}

Supplementary Video: The demo procedures of how to reproduce a real-world object or a virtual 3D model by SMNet.

\backmatter


\end{document}